\documentclass[letterpaper]{article} 
\usepackage{aaai2026}  
\usepackage{times}  
\usepackage{helvet}  
\usepackage{courier}  
\usepackage[hyphens]{url}  
\usepackage{graphicx} 
\usepackage{natbib}  
\usepackage{caption} 
\usepackage{algorithm}
\usepackage{algorithmic}
\usepackage{amsmath}
\usepackage{amsfonts}
\usepackage{booktabs} 
\usepackage{multirow}
\usepackage{bbding}
\usepackage{newfloat}
\usepackage{listings}
\DeclareCaptionStyle{ruled}{labelfont=normalfont,labelsep=colon,strut=off} 
\floatstyle{ruled}
\newfloat{listing}{tb}{lst}{}
\floatname{listing}{Listing}
\title{Multi-Agent Undercover Gaming: Hallucination Removal Through Counterfactual Test for Multimodal Reasoning}
\author{
    Dayong Liang\textsuperscript{\rm 1,2}\thanks{These authors contributed equally to this work.},
    Xiao-Yong Wei\textsuperscript{\rm 4,3,2}\footnotemark[1],
    Linhui Xiao\textsuperscript{\rm 2}\thanks{Corresponding author.},
    Changmeng Zheng\textsuperscript{\rm 3}\thanks{Corresponding author.}
}
\affiliations{
    \textsuperscript{\rm 1}School of Future Technology, South China University of Technology, Guangzhou, China\\
    \textsuperscript{\rm 2}PengCheng Laboratory, Shenzhen, China\\
    \textsuperscript{\rm 3}PolySmart, Department of Computing, The Hong Kong Polytechnic University, Hong Kong, China\\
    \textsuperscript{\rm 4}College of Computer Science, Sichuan University, Chengdu, China\\

    ft\_ldy@mail.scut.edu.cn, \{x1wei, changmeng.zheng\}@polyu.edu.hk
    
}

\usepackage{bibentry}

\begin{document}

\maketitle

\begin{abstract}
Hallucination continues to pose a major obstacle in the reasoning capabilities of large language models (LLMs).
Although the Multi-Agent Debate (MAD) paradigm offers a promising solution by promoting consensus among multiple agents to enhance reliability, it relies on the unrealistic assumption that all debaters are rational and reflective, which is a condition that may not hold when agents themselves are prone to hallucinations.
To address this gap, we introduce the Multi-agent Undercover Gaming (MUG) protocol, inspired by social deduction games like ``Who is Undercover?''. 
MUG reframes MAD as a process of detecting ``undercover'' agents (those suffering from hallucinations) by employing multimodal counterfactual tests. 
Specifically, we modify reference images to introduce counterfactual evidence and observe whether agents can accurately identify these changes, providing ground-truth for identifying hallucinating agents and enabling robust, crowd-powered multimodal reasoning.
MUG advances MAD protocols along three key dimensions: (1) enabling factual verification beyond statistical consensus through counterfactual testing; (2) introducing cross-evidence reasoning via dynamically modified evidence sources instead of relying on static inputs; and (3) fostering active reasoning, where agents engage in probing discussions rather than passively answering questions. 
Collectively, these innovations offer a more reliable and effective framework for multimodal reasoning in LLMs.
\end{abstract}

\begin{links}
    \link{Code}{https://github.com/YongLD/MUG.git}
    \link{Extended version}{https://arxiv.org/abs/2511.11182}
\end{links}

\section{Introduction}
Hallucination remains one of the most difficult challenges in LLM reasoning\cite{huang2025survey,liu2024survey,peng2025aligning}.
This stems from the fact that reasoning in these models is primarily constructed on token-level statistics (such as frequency and causality), often without explicit enforcement of factual correctness.
To tackle this problem, the Multi-Agent Debate (MAD) paradigm\cite{liang2024encouraging} has gained popularity, as it aims to generate a wider range of evidence, with the prevailing consensus among agents considered more trustworthy.
Although MAD has shown promising results, certain limitations persist.

Common MAD protocols are typically derived from human debate structures, such as majority voting, super judge, or blueprint debate.
However, there are fundamental differences between LLMs and human debaters.
While humans are capable of rational discussion, listening to others, reflecting, and updating their views as needed, it is still uncertain whether LLM debaters can do the same.
Consequently, protocols that rely heavily on debater rationality may not be optimal for MAD.
Therefore, there is a need for protocols that do not assume rationality and can effectively identify irrational debaters, particularly those susceptible to hallucinations.

\begin{figure*}
    \centering
    \includegraphics[width=6.8in]{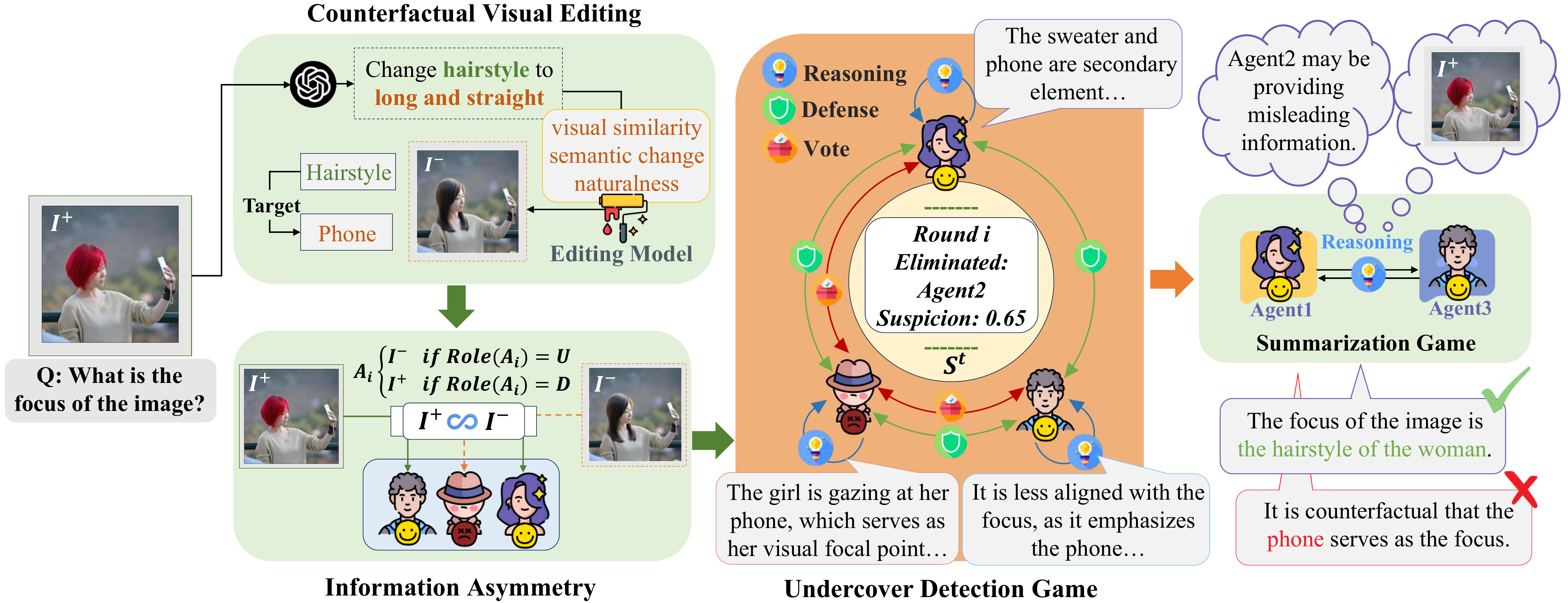}
    \caption{Overview of the Counterfactual Undercover Debate Framework, illustrating the dynamic interactions between agents, the current game state, and the decision-making process influenced by counterfactual information. The diagram highlights the roles of normal and undercover agents, as well as the flow of information and reasoning throughout the game.} 
    \label{model}
\end{figure*}

We draw inspiration from social deduction games like ``Who is Undercover?'', where all participants risk being the undercover and must debate to identify the undercover through factual or counterfactual challenges, often by posing further questions to each other.
This closely parallels the task of detecting hallucinating agents.
However, there is limited research on how to adapt such protocols for LLMs, especially regarding the design of factual and counterfactual tests when dealing with complex, multimodal evidence.
In this paper, we tackle these challenges by introducing the Multi-agent Undercover Gaming (MUG) protocol for MAD.
The core idea, illustrated in Figure~\ref{model}, is straightforward.
Given a question ``what is the focus of the image?'', we generate multimodal counterfactual tests by altering the reference image to include counterfactual evidence (\textit{e.g.}, changing a girl with red hair to a girl with natural black hair). In this scenario, the main focus of the image shifts from the \textit{hairstyle} to the \textit{phone}. We then observe whether each agent can correctly identify the modification. 
Because these changes are intentionally introduced, we have ``ground-truth'' to determine which agents are ``undercover'' (\textit{i.e.}, ones suffering from hallucinations). 
This allows us to harness the wisdom of the crowd for more robust and reliable multimodal reasoning.

MUG brings several key innovations to MAD protocols:
\begin{itemize}
    \item \textbf{Factual Verification vs. Statistical Consensus:} By employing counterfactual tests, MUG enables direct factual verification rather than relying solely on group consensus. 
    This addresses the limitation of LLMs, which often lack explicit fact-checking mechanisms during training.

    \item \textbf{Cross-Evidence vs. Single-Source:} By modifying images, MUG generates additional, dynamic sources of evidence for cross-referencing, as opposed to traditional MAD approaches that depend on a single, static source (such as the original question and reference image). 
    This not only facilitates verification but also increases the likelihood of detecting agents prone to hallucination.

    \item \textbf{Active vs. Passive Reasoning:} Existing protocols typically involve passive reasoning, where agents simply respond to a given question. 
    In contrast, MUG introduces counterfactual tests that require agents to actively engage by posing questions for discussion. 
    This approach better leverages the capabilities of LLMs and paves the way for developing more effective protocols.
    
\end{itemize}

\section{Related Work}
Early efforts in multimodal reasoning focused on aligning visual and textual representations, leading to the development of Multimodal LLMs (MLLMs) \cite{liu2024improved,ijcai2025p772} and the adoption of in-context learning (ICL) \cite{min2022rethinking} and chain-of-thought (CoT) reasoning \cite{wei2022chain}. These techniques empower models to decompose complex tasks into interpretable intermediate steps, enhancing transparency and performance. The evolution of CoT into multimodal contexts, termed Multimodal Chain-of-Thought (MCoT) reasoning, has produced a range of architectures, from linear \cite{wei2022chain, chongwahngo2008trecvid} to graph-based representations \cite{besta2024graph,yuan2023joint}, and specialized methods for different modalities, such as Multimodal-CoT \cite{zhang2024multimodal}, MVoT \cite{li2025imagine}, and Video-of-Thought \cite{fei2024video}. The paradigm of multi-agent systems (MASs) \cite{dorri2018multi} has gained traction as a means to overcome the intrinsic limitations of individual LLMs, such as hallucination and limited reasoning depth. MASs leverage the collective intelligence of multiple LLM-based agents, enabling distributed knowledge retention \cite{zheng2024picture}, long-term planning \cite{torreno2017cooperative}, and specialization through collaborative problem-solving \cite{lu2023chameleon,wang2025model}. However, existing MAD frameworks often converge prematurely, suppressing minority viewpoints and failing to simulate real-world adversarial reasoning conditions. Counterfactual reasoning, rooted in structural causal models and do-calculus \cite{gong2024distribution, chongwahngo2007trecvid}, is essential for understanding how outcomes change under hypothetical interventions. In AI, counterfactuals enhance interpretability and robustness by enabling models to generate alternative scenarios and assess decision-making under uncertainty \cite{guidotti2024counterfactual,wang2024survey}. Recent efforts have extended counterfactual reasoning to LLMs, with approaches ranging from commonsense-based scenario generation \cite{zhang2024learning,chatzi2025counterfactual} to graph-based causal inference using external tools, as exemplified by CausalCoT \cite{jin2023cladder} and CausalTool \cite{hua2024improving}. While agent-based frameworks like Causal Agent \cite{han2024causal} integrate LLMs with causal tools for intervention analysis, they often lack explicit counterfactual reasoning, limiting applicability in complex, high-dimensional settings. Our work distinguishes itself by embedding counterfactual learning directly into the multi-agent collaboration process, enhancing both the robustness and interpretability of multimodal reasoning. 

\begin{figure}
    \centering
    \includegraphics[width=\linewidth]{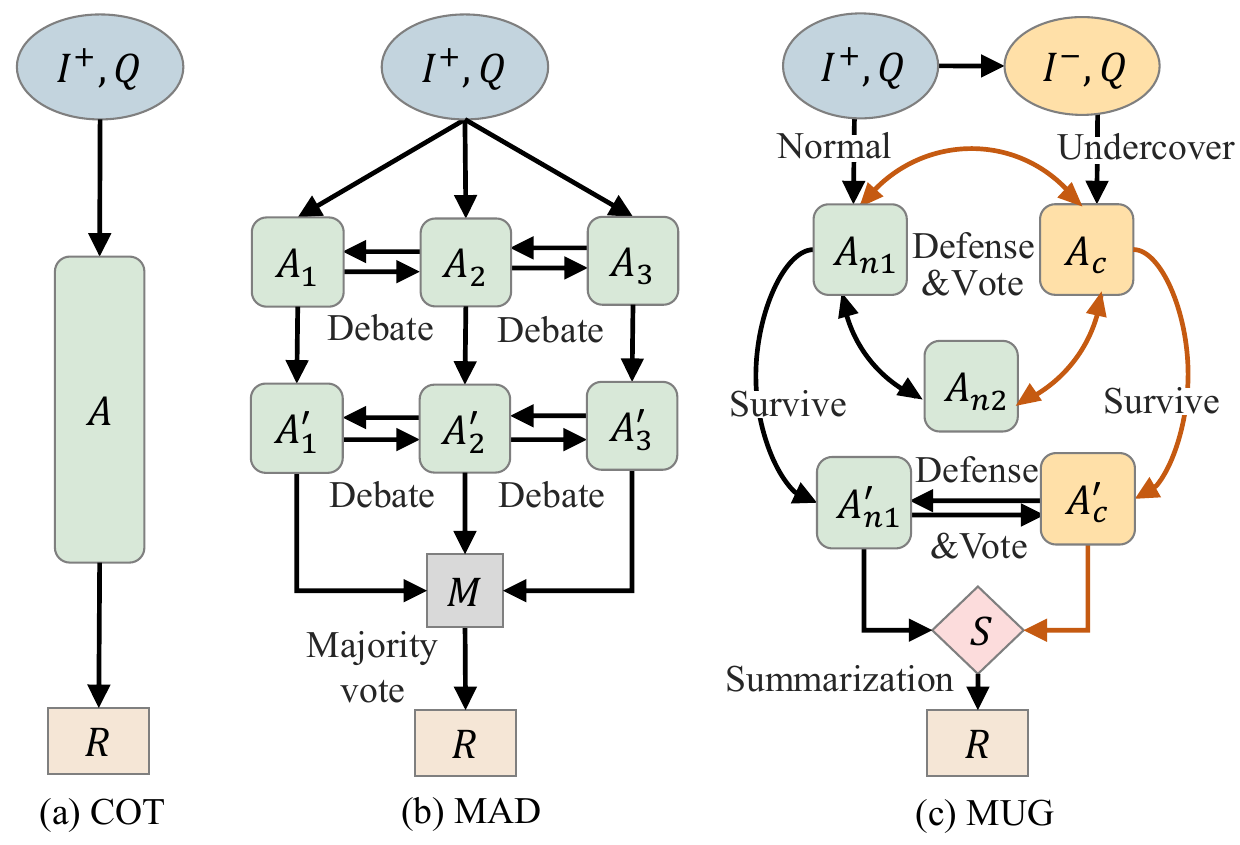}
    \caption{Comparison of Chain-of-thought (CoT), Multi-agent Debate (MAD), and our proposed Multi-agent Undercover Gaming (MUG). Q: input question, $I^+$: input image, $I^-$: edited counterfactual image, A: MLLM agent, R: output response.}
    \label{fig:mug}
\end{figure}

\section{Method}

\subsection{The Gaming System}
Multi-agent Undercover Gaming (MUG) takes place within a system $\mathcal{S}$, defined as a tuple of four components: a question $\mathcal{Q}$, a set of $N$ agents $\mathcal{A}=\{A_i\}_1^N$ who act as players or debaters, a collection of debating functions $\mathcal{F}$, and a set of responses $\mathcal{R}=\{{R}_i\}_1^N$ generated by the agents.
The system evolves dynamically over time, with its state at time $t$ denoted by $\mathcal{S}^t$.
At each time step $t$, the system receives the question $\mathcal{Q}$ as input. 
Each agent $A_i$ applies a function $f \in \mathcal{F}$ to engage in the debate, producing a response ${R}_i^t = f(\mathcal{Q}, A_i, t)$.
Formally, the state of the system at time $t$ is formulated as:
\begin{equation}
\mathcal{S}^t = (\mathcal{Q}, \mathcal{A}, \mathcal{F}, \mathcal{R}^t).
\end{equation}

The question is composed of a textual prompt $Q$, a factual reference image $I^+$, and a counterfactual image $I^-$, formally expressed as
\begin{equation}
    \mathcal{Q}=(Q,I^+,I^-),
\end{equation}
where generating the counterfactual image $I^-$ is a key aspect of MUG, to be discussed in detail later.

At the outset, one agent is assigned the counterfactual reference $I^-$ and takes on the role of the undercover agent, denoted as $Role(A_i) = U$. All other agents serve as regular debaters, with $Role(A_j) = D$.
The game proceeds with the agents participating in one of two modes:
\begin{itemize}
    \item \textbf{Undercover Detection Game} ($Game(\mathcal{S}^t)=D$): The objective is to identify the undercover agent.
    \item \textbf{Summarization Game} ($Game(\mathcal{S}^t)=M$): Agents work together to synthesize and provide a final answer to the question by combining their perspectives.
\end{itemize}
The game operates in detection mode until the undercover agent is identified. 
Once discovered, the undercover agent is removed from the game, and the system transitions to summarization mode to produce the final answer.

\subsection{Generation of Counterfactual Image $I^-$}
The construction of $I^-$ introduces asymmetric information by subtly altering specific details of the factual reference $I^+$.
The underlying idea is that these minor differences between the factual and counterfactual images will steer group discussions toward fine-grained semantic distinctions.
In contrast to existing multimodal reasoning methods \cite{liang2025seeing}, many of which rely heavily on image captioning and tend to settle on broad concepts (since captioning models are often trained on datasets like COCO, where labels are typically general), emphasizing discussion around subtle differences encourages participants to identify and reason about details directly relevant to the question, potentially leading to more precise and nuanced understanding.
To accomplish this, we begin with a cross-modal analysis to determine the type of edit and the target objects. 
Based on this analysis, we then prompt the image generator to perform the corresponding modifications.

\noindent\textbf{Edit Type and Targets Identification: } We leverage an LLM to classify the question $Q$ and map it to one of several predefined edit types.
For example, a ``How Many'' question is mapped to Quantity Editing, while a ``What Object'' question corresponds to Object Editing.
To pinpoint the specific targets for editing, we construct a scene graph of the factual image $I^+$, which allows for straightforward identification of relevant objects based on the chosen edit type.
Next, we prompt the LLM to generate an appropriate image editing prompt using the determined edit type and targets.
This workflow is illustrated in Figure~\ref{model}.
We observe that this approach does not heavily depend on LLM capabilities, as both question type recognition and target extraction are relatively straightforward tasks for modern language models.

\noindent\textbf{Modification for Counterfactual Reference:}
Following the detail-driven logic, the generation of $I^-$ must satisfy three key constraints:
\begin{itemize}
    \item Maximal Visual Similarity: $I^-$ should maximize visual similarity $C_{vs}=Sim^V(I^+, I^-)$ to $I^+$, helping the undercover agent remain concealed and preventing premature detection before critical scrutiny. Visual similarity is assessed by comparing ViT embeddings of both images.

    \item Semantic Consistency: The modification should retain the general semantics of the factual reference, maximizing semantic similarity. This is measured by the similarity between their CLIP embeddings, denoted as $C_{sc}=CLIP^S(I^+, I^-)$.

    \item Naturalness: $I^-$ should appear as a natural image, minimizing any artificial artifacts. This is quantified using the FID score, $C_{na}=FID(I^-)$.
\end{itemize}

\begin{figure}
    \centering
    \includegraphics[width=1.0\linewidth]{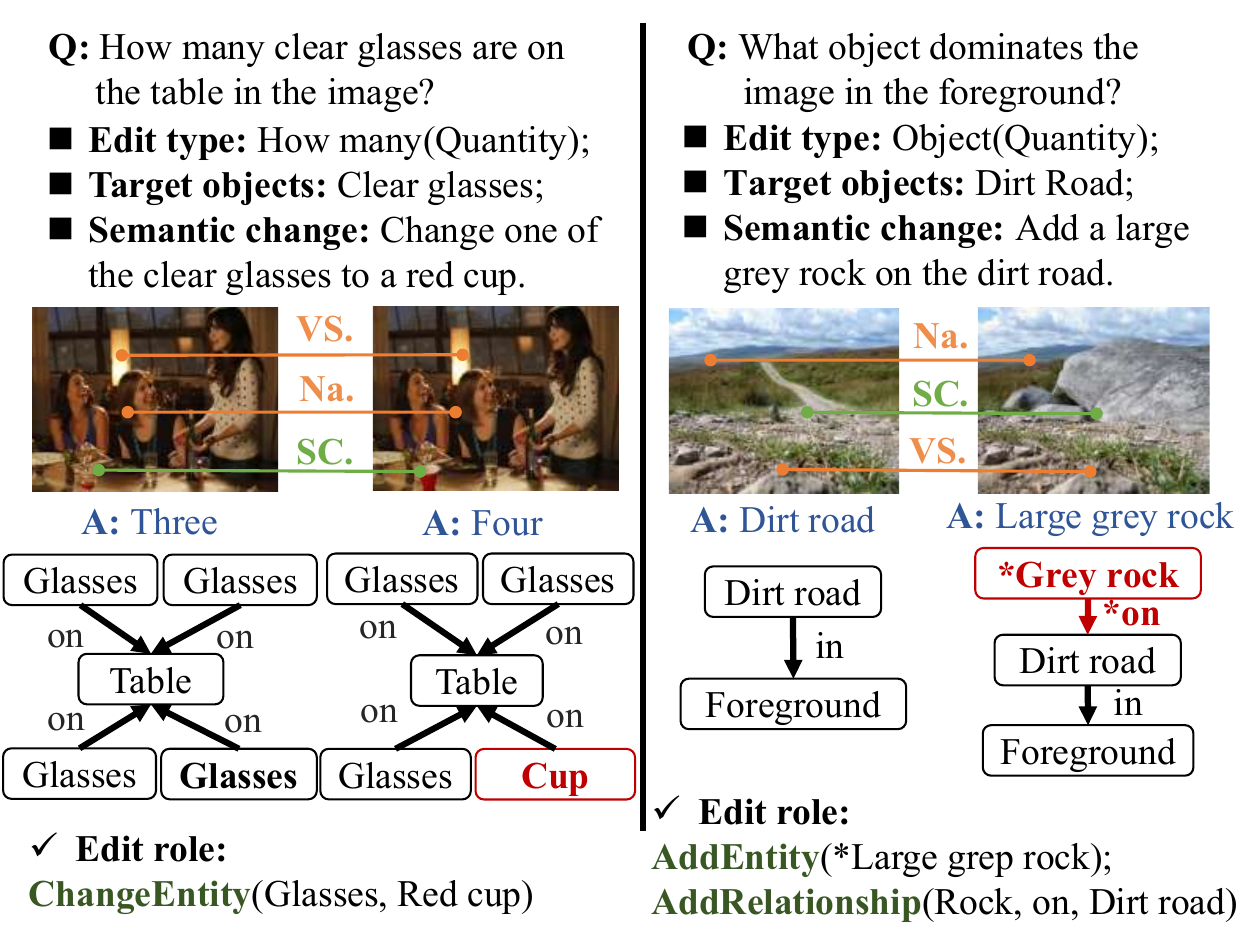}
    \caption{Illustration of counterfactual editing constraints, including visual similarity, semantic change, and naturalness. }
    \label{model1}
\end{figure}

For image generation, we utilize the Step1X-Edit \cite{liu2025step1x-edit} model. 
These constraints are enforced at the prompt level (see Appendix for prompt details) and are further validated in post-processing. 
The generated image is accepted only if the overall generation score meets a specified confidence threshold:
\begin{equation}
    \alpha\cdot C_{vs}+\beta\cdot C_{sc}+\gamma\cdot C_{na}\geq c
\end{equation}
Otherwise, the generation is repeated until these criteria are met.
The generation process is illustrated in Figure~\ref{model1}.

\subsection{Undercover Detection Game Mode}
After $I^-$ is generated, one agent will be assigned as the undercover and given $I^-$, while the others are assigned with $I^+$. The system enters the undercover detection mode.
The mode consists of two steps of reasoning and voting.
In reasoning, each agent is instructed to state its understanding of the question $Q$ with respect to the reference image assigned $I^+$ or $I^-$ in a faithful manner, while reduce its chance to be identified as the undercover. 

\noindent \textbf{Reasoning Phase:} In each round $t$, every agent $A_i$ generates a response $R_i^t$ that balances faithful reasoning with strategic concealment or detection objectives. The reasoning function for each agent incorporates their assigned image and role:
\begin{equation}
R_i^t = f_{reason}(Q, I_i, Role(A_i), \mathcal{H}^{t-1})
\end{equation}
where $I_i \in \{I^+, I^-\}$ represents the image assigned to agent $A_i$, and $\mathcal{H}^{t-1} = \{R_k^{t'}, V_k^{t'} | k \in \{1,...,N\}, t' < t\}$ captures the historical context from previous rounds, including all previous responses and voting decisions. For regular agents ($Role(A_i) = D$), the objective is to provide accurate reasoning based on $I^+$ while identifying inconsistencies that might reveal the undercover agent. For the undercover agent ($Role(A_u) = U$), the challenge is to generate plausible reasoning based on $I^-$ while avoiding detection.

The reasoning strategy for regular agents follows:
\begin{equation}
R_i^t = \arg\max_{r \in \mathcal{R}_i} \left[ Acc(r, Q, I^+) + \lambda \cdot Det(r, \mathcal{H}^{t-1}) \right]
\end{equation}
where $Acc(r, Q, I^+)$ measures the accuracy of reasoning $r$ with respect to the question and factual image, and $Det(r, \mathcal{H}^{t-1})$ captures the detection value of the reasoning given historical interactions.

For the undercover agent, the strategy balances plausibility with concealment:
\begin{equation}
R_u^t = \arg\max_{r \in \mathcal{R}_u} \left[ Pla(r, Q, I^-) - \mu \cdot Sus(r, \mathcal{H}^{t-1}) \right]
\end{equation}
where $Pla(r, Q, I^-)$ measures reasoning plausibility based on the counterfactual image, and $Sus(r, \mathcal{H}^{t-1})$ quantifies the suspicion it generates.

\noindent \textbf{Voting Phase:} Following the reasoning phase, agents engage in strategic voting to identify the undercover agent. Each agent $A_i$ casts a vote $V_i^t$ for the agent they suspect to be the undercover:
\begin{equation}
V_i^t = \arg\max_{j \neq i} \sum_{k=1}^{4} w_k \cdot \phi_k^{ij}(t)
\end{equation}

The voting decision incorporates four key factors derived from the reasoning analysis:

\begin{itemize}
    \item Inconsistency Score $\phi_1^{ij}(t)$: Measures logical inconsistencies in agent $j$'s reasoning compared to the expected reasoning based on $I^+$
    \item Deviation Score $\phi_2^{ij}(t)$: Quantifies how much agent $j$'s responses deviate from the consensus formed by other agents
    \item Detail Accuracy $\phi_3^{ij}(t)$: Assesses the accuracy of specific visual details mentioned by agent $j$
    \item Behavioral Suspicion $\phi_4^{ij}(t)$: Captures suspicious behavioral patterns such as vague responses or defensive language
\end{itemize}

The elimination decision is made by majority voting:
\begin{equation}
\hat{j} = \arg\max_{j} |\{i : V_i^t = j\}|
\end{equation}

\noindent \textbf{Round Progression:} The detection game continues iteratively until one of the termination conditions is met:
\begin{equation}
T^t = \begin{cases}
\text{True} & \text{if } Role(A_{\hat{j}}) = U \text{ (undercover found)} \\
\text{True} & \text{if } |\mathcal{A}^*| \leq 1 
\text{ (regular agents)} \\
\text{True} & \text{if } t > t_{max} \text{ (timeout)} \\
\text{False} & \text{otherwise}
\end{cases}
\end{equation}
where $ \mathcal{A}^*=\{A_i | Role(A_i) = D\}$. When the undercover agent is successfully identified and eliminated, the remaining agents $\{A_i | i \neq \hat{j}\}$ proceed to the summarization game mode with their shared understanding based on the factual image $I^+$.

\noindent \textbf{Strategic Dynamics:} The detection phase exploits information asymmetry to create strategic tension. Regular agents must balance detailed, accurate reasoning (which aids identification) against overly specific claims that may appear suspicious. The undercover agent must generate responses that align with the group while reasoning from different visual information. This dynamic promotes precise, detail-oriented reasoning across all agents, as vague responses become strategically disadvantageous. Regular agents demonstrate access to $I^+$ through specific details, while the undercover agent must carefully balance plausibility and accuracy to evade detection.

\begin{table*}
  \centering
  \small{
  \begin{tabular}{l|cc|cccc|cccc}
    \toprule
       \multirow{2}{*}{Model} & \multicolumn{1}{|c}{MMMU} & \multicolumn{1}{c|}{MMStar} & \multicolumn{4}{c|}{HallusionBench} & \multicolumn{4}{c}{POPE} \\
    \cmidrule(lr){2-3} \cmidrule(lr){4-7} \cmidrule(lr){8-11}
     & Acc. & Acc. & aAcc. & fAcc. & qAcc. & Avg. & Acc. & Pre. & Rec. & F1 \\
    \midrule

    DeepSeek-VL-7B & 38.3 & 40.5 & 53.9 & 24.9 & 24.6 & 34.5 & 86.8 & 94.3 & 78.4 & 85.6 \\
    LLaVA-OneVision-7B & 47.9 & 61.9 & 48.4 & 21.4 & 25.1 & 31.6 & 87.9 & 96.7 & 78.5 & 86.6 \\
    LLaVA-Next-Llama3-8B  & 43.1 & 43.9 & 52.3 & 25.7 & 21.3 & 33.1 & 87.1 & 88.1 & 95.1 & 87.1 \\
    InternVL2-8B  & 50.2 & 61.5 & 63.9 & 35.0 & 36.0 & 45.0 & 86.0 & 96.6 & 74.7 & 84.2 \\
    Gemini-1.5-Pro & 60.6 & 59.1 & 63.0 & 36.1 & 37.6 & 45.6 & 88.6 & 93 & 83.9 & 88.2 \\
    GPT4o\_mini (detail-high)  & 57.3 & 46.5 & 61.9 & 41.3 & 34.9 & 46.1 & 84.2 & 95.5 & 71.7 & 81.9 \\
    GPT-4v (detail-high) & 53.8 & 43.9 & 65.8 & 38.4 & 35.2 & 46.5 & 83.9 & 93.9 & 72.5 & 81.8 \\
    Claude3.5-Sonnet & 65.9 & 62.2 & 66.4 & 41.6 & 41.8 & 49.9 & 78.7 & 97.1 & 59.2 & 73.6 \\
    \midrule
    Qwen2.5VL-7B  & 45.0 & 61.2 & 64.8 & 34.9 & 39.6 & 46.4 & 87.4 & \underline{\textbf{96.8}} & 77.3 & 85.9 \\
    Qwen2.5VL-7B (Self-Refine)  & 45.8 & 61.5 & 67.3 & 38.2 & 40.9 & 48.8 & 85.9 & 97.4 & 73.8 & 84.0 \\
    Qwen2.5VL-7B (MAD-Vote)  & 44.7 & 57.4 & 56.4 & 26.9 & 30.1 & 37.8 & 80.0 & 80.4 & 68.7 & 74.1 \\
    Qwen2.5VL-7B (MAD-Judge)  & 47.4 & 62.3 & 64.5 & 43.2 & 42.9 & 50.2 & 85.2 & 94.4 & 74.9 & 83.5 \\
    Qwen2.5VL-7B (MUG(\textbf{Ours}))  & \underline{\textbf{50.3}}  & \underline{\textbf{63.8}} & \underline{\textbf{69.4}} & \underline{\textbf{43.9}} & \underline{\textbf{47.9}} & \underline{\textbf{53.8}} & \underline{\textbf{88.4}} & 95.6 & \underline{\textbf{80.5}} & \underline{\textbf{87.4}} \\
    \midrule
    InternVL3-14B  & 59.8 & 68.7 & 69.8 & 47.7 & 47.7 & 55.1 & 89.3 & 92.1 & 86.5 & 89.5 \\
    InternVL3-14B (Self-Refine)  & 45.9 & 61.2 & 70.3 & 46.5 & 45.5 & 54.1 & 88.6 & 91.9 & 86.7 & 89.2 \\
    InternVL3-14B (MAD-Vote) & 55.2 & 62.9 & 70.5 & 48.6 & 46.4 & 55.2 & 89.4 & \underline{\textbf{94.2}} & 83.9 & 88.8 \\
    InternVL3-14B (MAD-Judge)  & 60.2 & 68.9 & 72.5 & 47.9 & 48.9 & 56.4 & 89.6 & 94.1 & 87.3 & 90.6 \\
    InternVL3-14B (MUG(\textbf{Ours})) & \underline{\textbf{60.7}}  & \underline{\textbf{69.1}} & \underline{\textbf{73.3}} & \underline{\textbf{51.2}} & \underline{\textbf{49.5}} & \underline{\textbf{58.0}} & \underline{\textbf{90.1}} & 94.1 & \underline{\textbf{88.2}} & \underline{\textbf{91.1}} \\
    \bottomrule
  \end{tabular}
  }
  \caption{Performance comparison across multiple benchmarks. The improvements against  baselines are statistically significant by one-tailed paired $t$-test with $p$ $<$ 0.01. 
  Acc.: accuracy, aAcc.: average accuracy, fAcc.: unique figure accuracy, qAcc.: unique quetion accuracy, Avg.: average score of the three metrics. Pre.: precision, Rec.: recall.}
  \label{tab:multimodel}
\end{table*}

\subsection{Summarization Game Mode}

Upon successful identification and elimination of the undercover agent, the system transitions to summarization mode with the objective $Game(\mathcal{S}^t) = M$. In this phase, the remaining agents collaborate to synthesize their individual understandings and produce a final answer to the question $Q$.

\noindent \textbf{Collaborative Reasoning:} The summarization phase operates through structured dialogue where agents build upon each other's insights. Each remaining agent $A_i$ contributes to the collective reasoning process:
\begin{equation}
R_i^{sum} = f_{sum}(Q, I^+, \mathcal{R}_{collect}^{t-1}, \mathcal{H}^{t-1})
\end{equation}
where $\mathcal{R}_{collect}^{t-1}=\{R_k^{t'} | k \neq \hat{j}, t' < t\}$ represents the accumulated collective reasoning from previous summarization rounds.

\noindent \textbf{Final Answer Generation:} The collaborative process culminates in the generation of a final answer that synthesizes the collective intelligence of the remaining agents:\begin{equation}
R^{answer} = f_{answer}(R^{sum}, Q, I^+)
\end{equation}

The summarization mode terminates when agents reach sufficient consensus or when the maximum number of summarization rounds is exceeded, ensuring that the final answer represents the best collective reasoning based on the factual visual information.
The function collection $\mathcal{F}$ encompasses all debating functions used throughout the system: $\mathcal{F} = \{f_{reason}, f_{sum}, f_{answer}\}$.

\section{Experiments}

\subsection{Experimental Setup}

We conduct comprehensive experiments to evaluate the effectiveness of our Multi-agent Undercover Gaming (MUG) framework across multiple dimensions: multimodal reasoning accuracy, hallucination detection capability, and agent reliability identification. Our experimental design systematically validates each component of the proposed framework while demonstrating its superiority over existing approaches. Implementation details can be found in Appendix.

\subsubsection{Datasets}

We evaluate our framework on four datasets covering different aspects of multimodal reasoning:

\noindent \textbf{General Reasoning Datasets:} MMStar \cite{chen2024we} and MMMU \cite{yue2024mmmu} serve as comprehensive benchmarks for evaluating overall multimodal reasoning capabilities. MMStar contains 1,500 challenging visual questions spanning multiple domains, while MMMU includes 11,500 college-level multimodal problems requiring sophisticated reasoning across disciplines.

\noindent \textbf{Hallucination Detection Datasets:} HallusionBench \cite{guan2024hallusionbench} and POPE \cite{Li-hallucination-2023} focus on hallucination detection in multimodal systems. HallusionBench provides 346 samples to induce hallucinations through misleading visuals, while POPE offers 3,000 yes/no questions to assess object hallucination in image descriptions.

\subsubsection{Baselines}
We evaluate our MUG framework against several baseline models to benchmark its performance across multimodal reasoning tasks. The baseline methods include: (1) \textbf{Single-Agent Baselines}, which can be categorized into open-source and closed-source models. The open-source models consist of DeepSeek-VL-7B \cite{lu2024deepseek}, LLaVA-OneVision-7B \cite{li2024llava}, LLaVA-NEXT-Llama3-8B \cite{llava-next}, InternVL2-8B\cite{chen2024internvl}, InternVL3-14B\cite{zhu2025internvl3} and Qwen2.5VL \cite{bai2025qwen2}. The closed-source models include Gemini-1.5-pro \cite{team2024gemini}, GPT4O\_mini \cite{hurst2024gpt}, GPT4v \cite{gpt4v}, and Claude-3.5-Sonnet \cite{claude-3.5}. These models combine visual and textual information for reasoning tasks, using diverse techniques to improve multimodal understanding and accuracy across benchmarks. (2) \textbf{Self-Refine Framework} \cite{madaan2023self} is an approach to improve initial outputs from LLM through iterative feedback. (3) \textbf{ Multi-Agent Debate (MAD)} \cite{liang2024encouraging} is a classic approach where multiple agents debate and reach conclusions through majority voting or a super judge, directly comparing to our multi-agent framework.

\begin{figure}
    \centering
    \includegraphics[width=1.0\linewidth]{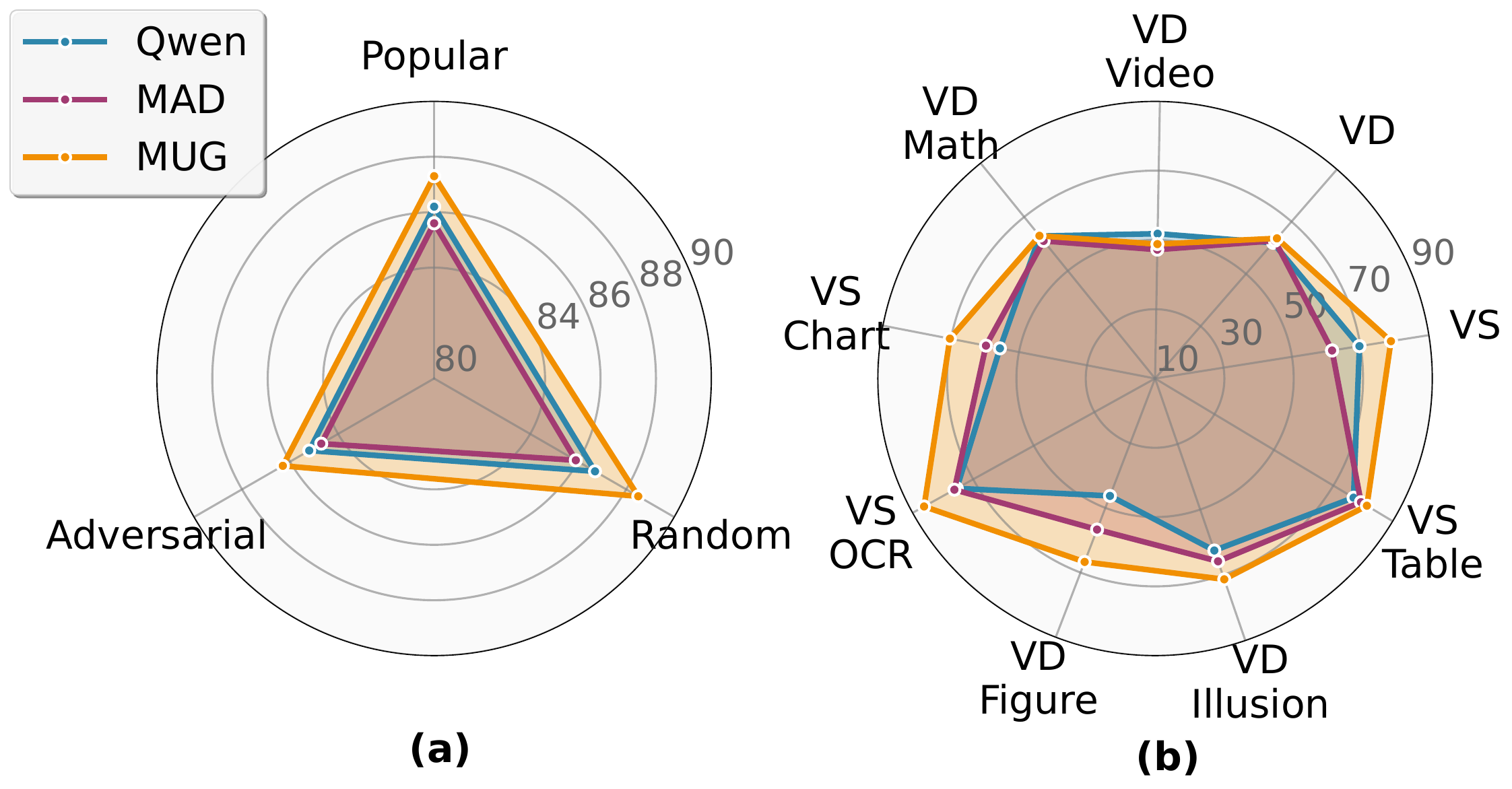}
    \caption{Performance comparison for different models across hallucination categories in the POPE and HallusionBench datasets.}
    \label{fig:radar_MUG}
\end{figure}

\subsection{Performance Comparison to SOTA Methods}
In this section, we present the main experimental results obtained from our MUG framework compared to several SOTA methods on four widely-used multimodal reasoning benchmarks. Overall, the integration of MUG has resulted in a significant improvement across different backbones. On the MMMU\_VAL benchmark, Qwen2.5VL-7B (MUG) achieves an accuracy of 50.3\%, showing a 5.3\% improvement over the baseline accuracy of 45.0\%. Meanwhile, InternVL3-14B (MUG) reaches 60.7\%, surpassing its baseline of 59.8\% by 0.9\%. In terms of the MMStar benchmark, Qwen2.5VL-7B (MUG) achieves an accuracy of 63.8\%, while InternVL3-14B (MUG) achieves 69.1\%, demonstrating the robustness of our framework across multiple evaluation metrics. Other observations that indicate MUG's advantage over SOTA methods include:

\noindent \textbf{MUG helps reduce the performance gap between large and small models.} Larger models typically outperform smaller ones. This observation holds true since we can find that the closed-source models like Gemini-Pro, GPT-4v and Claude-Sonnet outperform many open-source baselines including Qwen2.5VL and InternVL3. However, the introduction of MUG has led to a reduction in the performance gap between these two types of models. This can be seen in the Qwen2.5VL-7B model, which has reached the overall accuracy of 63.8\%, 53.8\% and 88.4\%, even better than the GPT-4v and Claude3.5-Sonnet models.

\noindent \textbf{MUG reinforces the multiagent collaboration.} Using Qwen2.5VL-7B as the base model, MUG achieves substantial improvements over traditional collaboration methods, outperforming Multi-Agent Debate with voting (MAD-Vote) by 5.6 points on MMMU (50.3\% vs 44.7\%) and 16.0 points on HallusionBench average score (53.8\% vs 37.8\%), while surpassing the more sophisticated MAD-Judge approach by 2.9 points on MMMU and 3.6 points on HallusionBench average. Similar performance patterns emerge with the larger InternVL3-14B model, where MUG surpasses MAD-Vote by 5.5 points on MMMU and achieves the highest scores across all HallusionBench metrics.

\begin{figure}
    \centering
    \includegraphics[width=\linewidth]{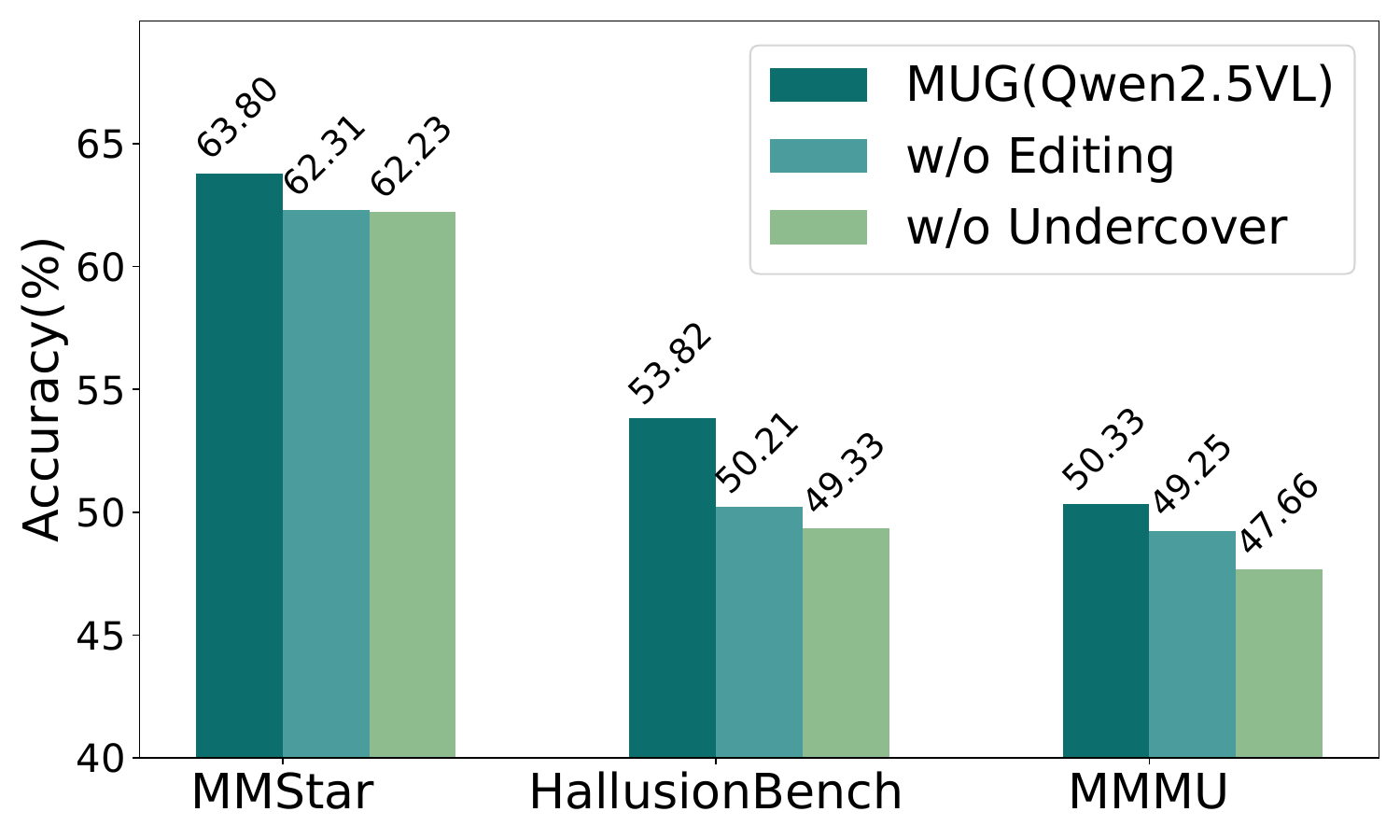}
    \caption{Ablation results demonstrating the impact of counterfactual editing and undercover mechanisms on performance across three benchmarks.}
    \label{fig:ablation_effects}
\end{figure}

\subsection{Performance on Hallucination Issues}
MUG demonstrates superior hallucination detection capabilities across both model architectures, achieving the highest average scores of 53.8\% (Qwen2.5VL-7B) and 58.0\% (InternVL3-14B) compared to all baseline methods. The framework shows particularly strong performance in figure accuracy, with improvements of 9.0 points over the base Qwen model (43.9\% vs 34.9\%) and 3.5 points over InternVL3 (51.2\% vs 47.7\%), indicating enhanced ability to detect visual inconsistencies. As illustrasted in Figure \ref{fig:radar_MUG}, category-specific analysis reveals MUG's strengths in visual similarity tasks (78.9\% vs 69.7\% baseline), OCR challenges (86.2\% vs 75.4\%), and figure understanding (66.7\% vs 46.3\%), while maintaining competitive performance across mathematical reasoning and video analysis tasks. The consistent improvements across different hallucination types demonstrate MUG's robust counterfactual reasoning mechanism effectively exposes model vulnerabilities.

On the POPE dataset, MUG achieves the highest overall accuracy scores of 88.4\% (Qwen2.5VL-7B) and 90.1\% (InternVL3-14B), outperforming traditional multi-agent approaches and single-model baselines. The framework shows balanced precision-recall trade-offs, with notable recall improvements of 3.2 points (80.5\% vs 77.3\%) for Qwen and 1.7 points (88.2\% vs 86.5\%) for InternVL3, indicating better detection of actual objects while maintaining high precision. Cross-category analysis on POPE reveals MUG's robustness across different evaluation scenarios, with the most significant gains in the random setting (88.5\% vs 86.7\%) and consistent improvements in adversarial (86.3\% vs 85.2\%) and popular (87.3\% vs 86.2\%) configurations, demonstrating the framework's effectiveness in mitigating various types of object hallucinations through strategic information asymmetry.

\subsection{Ablation Studies}

The ablation study results demonstrate that both core components of MUG contribute significantly to performance across all evaluation benchmarks. As shown in Figure \ref{fig:ablation_effects}, removing the counterfactual editing module leads to performance drops of 1.49 points on MMStar, 3.61 points on HallusionBench, and 1.08 points on MMMU, indicating that counterfactual visual modifications are crucial for creating effective reasoning conflicts. The elimination of the undercover agent mechanism results in more substantial degradations, with decreases of 1.57 points on MMStar, 4.49 points on HallusionBench, and 2.67 points on MMMU. The larger performance drops when removing the undercover agent, particularly on hallucination-focused benchmarks like HallusionBench, underscore the importance of the strategic gaming dynamics in exposing reasoning inconsistencies and enhancing collaborative verification processes.


\begin{table}
  \centering
  \small
  \begin{tabular}{lccc}
    \toprule
    Round & HallusionBench & MMMU & MMStar \\
    \midrule
    0 & 67.31 & 47.88 & 61.93 \\
    1 & \textbf{69.40} & \textbf{50.33} & 63.80 \\
    2 & 65.89 & 48.02 & \textbf{63.92} \\
    3 & 66.95 & 47.56 & 62.76 \\
    \bottomrule
  \end{tabular}
  \caption{Model performance with respect to the iteration round of game.}
  \label{tab:round_performance}
\end{table}

\subsection{Monitoring the Game Progress}
\subsubsection{Impact of Game Rounds}
We analyze the effect of the number of game rounds, which corresponds to the observation rounds set during the exchange of opinions before voting begins. The performance metrics across different rounds are shown in Table \ref{tab:round_performance}. The results indicate that the initial rounds yield higher performance, likely due to the freshness of opinions and the strategic dynamics of the debate. Specifically, the performance peaks at round 1, with a notable accuracy increase, reflecting the effectiveness of initial interactions among agents. As rounds progress, the performance stabilizes or slightly declines, suggesting diminishing returns on additional rounds.

\begin{figure}
    \centering
    \includegraphics[width=\linewidth]{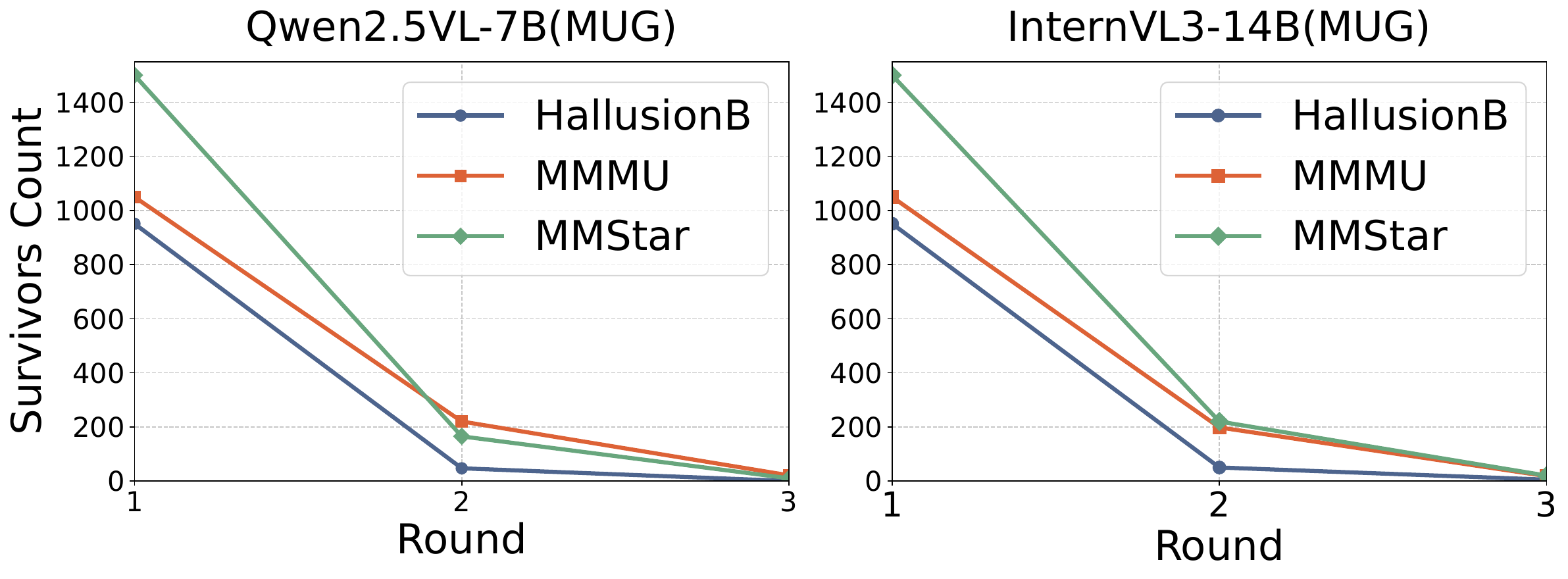}
    \caption{Statistics of survived undercover agents across three game rounds.}
    \label{fig:coun_vote}
\end{figure}

\subsubsection{Analysis on Game Termination}

Figure \ref{fig:coun_vote} illustrates the number of survived undercover agents between and within rounds. Different from traditional MAD methods which rely on a super judge or maximum setting of round to terminate debate, one strength of our proposed MUG framework is that we can easily know the game ends when the undercover is voted out. An interesting finding is that the sharp decline in the second round, which suggests that agents are effectively identifying and eliminating undercover agents based on the counterfactual evidence and reasoning processes employed in this dataset. In contrast,  the third or fourth round exhibits slower rates of elimination, with survival percentages hovering around 5\%. These results imply that the agents involving in these data samples are either less effective at identifying hallucinations or that the undercover agents are employing strategies that allow them to blend in more successfully within these specific scenarios.

\begin{figure}
    \centering
    \includegraphics[width=\linewidth]{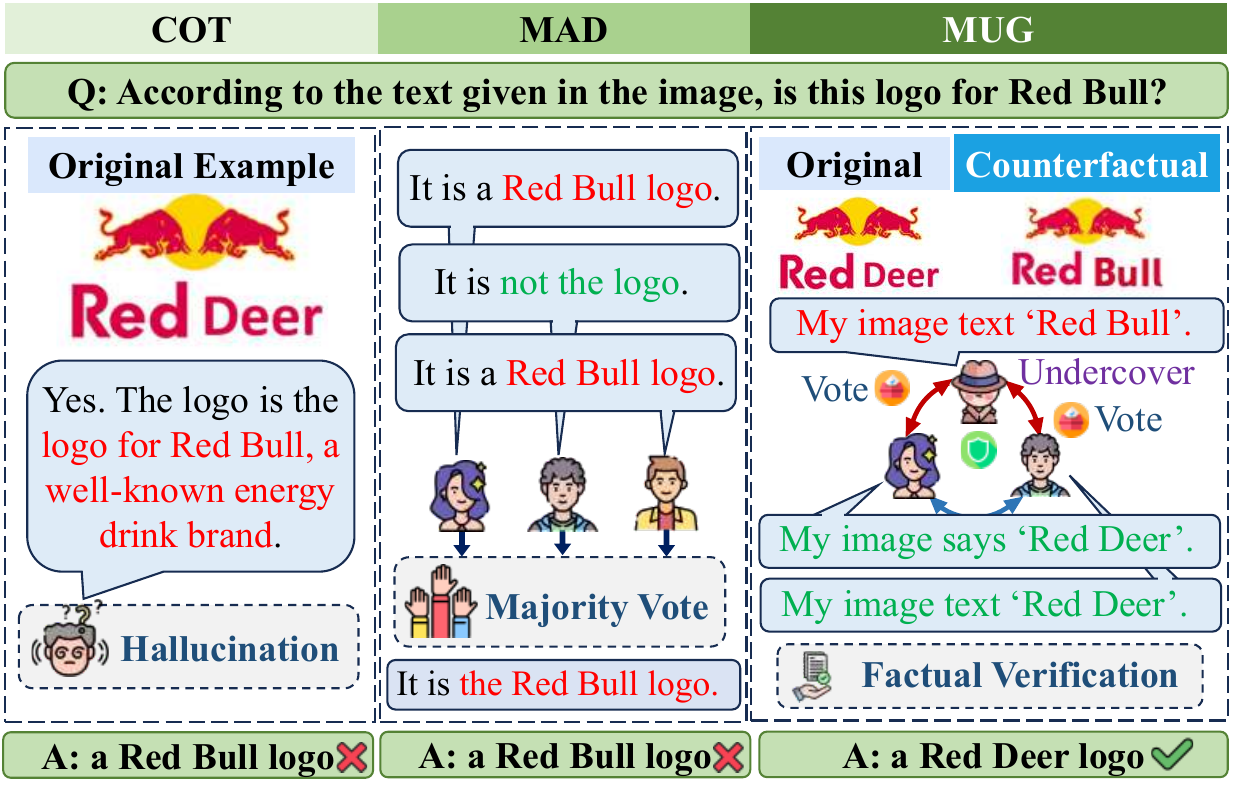}
    \caption{Case study of our proposed MUG framework compared with traditional chain-of-thought reasoning (CoT) and multiagent debate (MAD).}
    \label{fig:case}
\end{figure}

\subsection{Case Study}
Figure \ref{fig:case} presents a compelling case study highlighting the effectiveness of the Multi-agent Undercover Gaming (MUG) protocol in addressing hallucinations in multimodal reasoning scenarios. The left panel shows a traditional agent incorrectly identifying a logo reading ``Red Deer" as affiliated with Red Bull, while the middle panel illustrates Multi-Agent Debate (MAD) failing to resolve underlying inaccuracies through majority voting based on potentially flawed interpretations. In contrast, the right panel demonstrates how MUG introduces counterfactual evidence that prompts deeper analysis, with agents engaging in critical discussions around discrepancies such as conflicting phrases ``my image says `Red Bull'" versus ``Red Deer," ultimately enabling more accurate logo identification through collaborative verification mechanisms that surpass traditional debate approaches.

\section{Conclusion}

This paper presents the Multi-agent Undercover Gaming (MUG) protocol as a novel solution to address hallucination challenges in large language models (LLMs). By integrating counterfactual tests inspired by social deduction games, MUG enhances the traditional Multi-Agent Debate (MAD) framework, allowing agents to engage in active reasoning and verification. Our experimental results demonstrate that MUG significantly outperforms baseline models in multimodal reasoning accuracy and hallucination detection.

\section{Acknowledgments}
This work has been supported by National Natural Science Foundation of China (Grant No. 62372314) and Hong Kong Research Grants Council under the General Research Fund (project no. PolyU 15200023). The experimental part of this work was supported by the Centre for Large AI Models (CLAIM) of The Hong Kong Polytechnic University.

\bibliography{aaai2026}

\end{document}